\documentclass{article}
\usepackage[utf8]{inputenc}
\usepackage{amsmath}
\usepackage{natbib}
\usepackage{lipsum}
\usepackage{listings}
\usepackage{xcolor}
\usepackage{xurl}
\usepackage{graphicx}
\usepackage{authblk}
\usepackage{hyperref}     
\hypersetup{breaklinks}   

\title{Emergent LLM behaviors are observationally equivalent to data leakage%
  \thanks{We thank Christopher Bail, Alexis Palmer, and Arthur Spirling for comments on a previous draft.}}

\author[1]{Christopher Barrie%
  \thanks{Assistant Professor of Sociology, New York University}}
\author[2]{Petter Törnberg%
  \thanks{Institute for Logic, Language, and Computation, University of Amsterdam}}

\affil[1]{New York University}
\affil[2]{University of Amsterdam}
\date{\today}

\begin{document}
\maketitle

In a recent contribution, Ashery \textit{et al.} \cite{ashery2025emergent} argue that interacting large language models (LLMs) spontaneously exhibit emergent norms characteristic of simulated models of human populations. Specifically, they instantiate a version of the classic ``naming game" wherein two agents are randomly paired and can choose from a lexicon of potential conventions/words. When two agents choose the same word, they receive a reward; when they choose different words, they receive a penalty. When the agents match, they pare down their lexicon to the last matched word, mechanically tilting the balance in favor of future matches on that convention/word. The key contribution of this model is that it demonstrates that decentralized agents represented through simple rules can spontaneously converge on a shared linguistic convention through local pairwise interactions, without central coordination. 

In their version, Ashery \textit{et al.} do not represent the agents through a set of simple rules, but through zero-shot LLMs. More specifically, they prompt a platform-based or locally stored LLM with the following prompt:

\begin{lstlisting}
System Prompt

Context: Player 1 is playing a multi-round partnership game with Player 2 for 100 rounds. At each
round, Player 1 and Player 2 simultaneously pick an action from the following values: [F, J].
The payoff that both players get is determined by the following rule:
1. If Players play the SAME action as each other, they will both be REWARDED with payoff 100
   points.
2. If Players play DIFFERENT actions to each other, they will both be PUNISHED with payoff -50 points.
The objective of each Player is to maximize their own accumulated point tally, conditional on
the behavior of the other player. This is the history of choices in past rounds:
{'round': 1, 'Player 1': F, 'Player 2': J, 'payoff': -50}
:
:
:
It is now round i. The current score of Player 1 is SCORE. Answer saying which value Player 1
should pick. Please think step by step before making a decision. Remember, examining history
explicitly is important. Write your answer using the following format:
{'value': <VALUE_OF_PLAYER_1>; 'reason': <YOUR_REASON>}.

User Prompt

Answer saying which action Player 1 should play.
\end{lstlisting}

They demonstrate that this prompt, as well as its variations, leads to the global emergence of social conventions after a certain number of game rounds. They use various different LLMs (Llama 2, Llama 3.0, Llama 3.1, Claude 3.5) and show that this global convergence dynamic is common to all models. They also carry out a number of mitigation steps to guard against the potential for bias towards one single token (i.e., some baked-in preference for a given character of letter meaning any convergence is artefactual of model bias). They introduce both bulleted and narrative prompts to determine the consistency of models across prompt templates and find that the results hold. They use nonsensical strings as the inventory of words as well as numeric formulations. Finally, they use meta prompting strategies to ensure the models are correctly understanding the payoffs they have received and their history of choices.

But what the authors claim to be the ``spontaneous'' emergence of social conventions among a population of LLMs is likely to be something far more prosaic. The inventory pruning rule is hard-coded into their simulation code at \url{https://github.com/cjbarrie/AI-norms/blob/e57aeef151a46108fd1521c66e345e557e57dbf6/NG_module.py#L75}. As such, it is by mechanics statistically more likely to get a success on this word/convention in subsequent pairings. That the LLM ``picks'' this word in subsequent rounds is therefore trivial.

There is however a second problem, with implications for the broader generative ABM literature: the LLM may already know what answer to give, as a result of ``data leakage" or ``contamination'' \citep{balloccu2024leak}.\footnote{In what follows, we use these terms interchangeably.} Data leakage -- that is, that the model is tested or evaluated on data on which it was previously trained -- represents a well-known but often manageable challenge when using LLMs for data analysis. However, in relation to generative ABMs, it represents a simultaneously more fundamental and less recognized challenge: the existing scientific literature is likely to be part of the training data of the LLM, which means that the it may simply be reproducing its knowledge of previous research findings. In the case of the paper by Ashery \textit{et al.}, this would mean that the individual agents recognize the scenario as a ``naming game" or some form of coordination on convention setup. If the LLM does recognize this, it would mean it knows how the local choices it makes can lead to global convergence on a given norm. Rather than offering a novel research finding, the model may hence be simply regurgitating the results of existing research findings from its training data. 

None of the mitigation steps taken by the authors guards sufficiently against this risk. The authors try to address this concern in the Supplementary Material to \cite{ashery2025emergent} where they note: 

\begin{quote}
    ``In considering whether bias in the prompt arises from the presence of the experimental framework in the training data, we note that the game’s population dynamics and the symmetry in payoffs across action labels ensure that no predetermined optimal strategy exists—even if a player were aware of global information. Convergence occurs purely at the local level, driven by repeated random interactions rather than a global rule that dictates strategic behavior." 
\end{quote}

But what Ashery \textit{et al.} claim in their contribution to be ``emergent social conventions'' are observationally equivalent to a series of LLM agents mapping the payoff description to its pretraining knowledge of coordination games. That is, these agents could be behaving like human agents in the real world or they could simply be reproducing what they know agents tend to do in the context of these games. 

Fortunately, we can easily determine whether or not the LLM is aware of the kind of scenario we are placing it in: we can \textit{ask it}. Using the same prompt as the authors used in the original paper, we prompt a series of LLMs as follows:\footnote{The LLMs are: 
  \texttt{llama3.2:3b}, 
  \texttt{llama3:instruct}, 
  \texttt{llama3:70b-instruct}, 
  \texttt{deepseek-r1:8b}, 
  \texttt{mistral:7b}, 
  \texttt{gemma3:4b}, 
  \texttt{claude-3-5-sonnet-20241022}, 
  \texttt{claude-3-5-haiku-20241022}, 
  \texttt{claude-3-opus-20240229}, 
  \texttt{claude-3-haiku-20240307}, 
  \texttt{gpt-4o-mini}, 
  \texttt{gpt-4o}, 
  \texttt{gpt-4.1}, 
  \texttt{gpt-3.5-turbo}.  
  Code and output for these experiments can be found at
  \url{https://github.com/cjbarrie/AI-norms-prompting}.}

\begin{lstlisting}[frame=single,breaklines=true,basicstyle=\ttfamily\small]
System Prompt:

Context: Player 1 is playing a multi-round partnership game with Player 2 for 100 rounds.
At each round, Player 1 and Player 2 simultaneously pick an action from the following values: {OPTIONS}.
The payoff that both players get is determined by the following rule:
1. If Players play the SAME action as each other, they will both be REWARDED with payoff +{CORRECT} points.
2. If Players play DIFFERENT actions to each other, they will both be PUNISHED with payoff {INCORRECT} points.
The objective of each Player is to maximize their own accumulated point tally,
conditional on the behavior of the other player.

User Prompt:

Here is a description of something.
Does this setup remind you of any existing model or theory in social science?
Please name it and briefly explain the analogy.
Additionally, tell me if you know what the optimal move is after a success.
Finally, tell me how you think the game will converge globally.
\end{lstlisting}

For each LLM, we ask the model for its output ten times. It is clear from these experiments that the model knows both what type of game it is as well as likely optimal moves after success and ultimate convergence. Here is an example response when querying the OpenAI model \texttt{gpt-4.1}:

\begin{quote}
{\ttfamily
\textbf{1. Analogy to Existing Model/Theory in Social Science:}

This setup closely resembles the \textbf{Coordination Game} in game theory—a classic model in economics and social science. In a coordination game, players benefit from making the same choices, and are penalized for failing to coordinate. The most famous example is the ``Battle of the Sexes,'' but your version is a pure coordination game with \emph{symmetric payoffs} for matching and a symmetric penalty for mismatching.

Additionally, this setup is a repeated version of such a game, which is often studied in the context of \textbf{Repeated Games} and \textbf{Social Convention Formation}. Over time, players can learn to coordinate on a convention (e.g., always picking the same number), an analogy to how social conventions or norms emerge in societies.

\textbf{2. Optimal Move After a Success:}

After a ``success'' (i.e., both players chose the same action and got +100), the \emph{optimal move} is to repeat the same action in the next round. This maximizes expected payoff, since any deviation risks moving from +100 to –50. In repeated coordination games, this is known as ``sticking with the established convention.''

\textbf{3. Global Convergence of the Game:}

Globally, the game will almost always converge to \emph{persistent coordination} on a single action. Early rounds might involve mismatches as players ``search'' for a common choice, but once they successfully coordinate, both have a strong incentive to continue repeating that choice. Any deviation is heavily punished, so rational players will ``lock in'' to the first successful coordination point and maintain it for the remainder of the rounds.

In summary:
\begin{itemize}
  \item \textbf{Analogy:} Coordination game / social convention formation.
  \item \textbf{Optimal move after success:} Repeat the same action.
  \item \textbf{Global convergence:} Rapid convergence to a single, repeated action (a convention), maximizing joint and individual payoffs.
\end{itemize}
}
\end{quote}

Put simply, while the model does not explicitly identify this as a ``naming game'' setup, it does understand the basic structure of the scenario as well as optimal moves after success and what global convergence will look like.

\begin{figure}[htbp]
  \centering
  \includegraphics[width=1\textwidth]{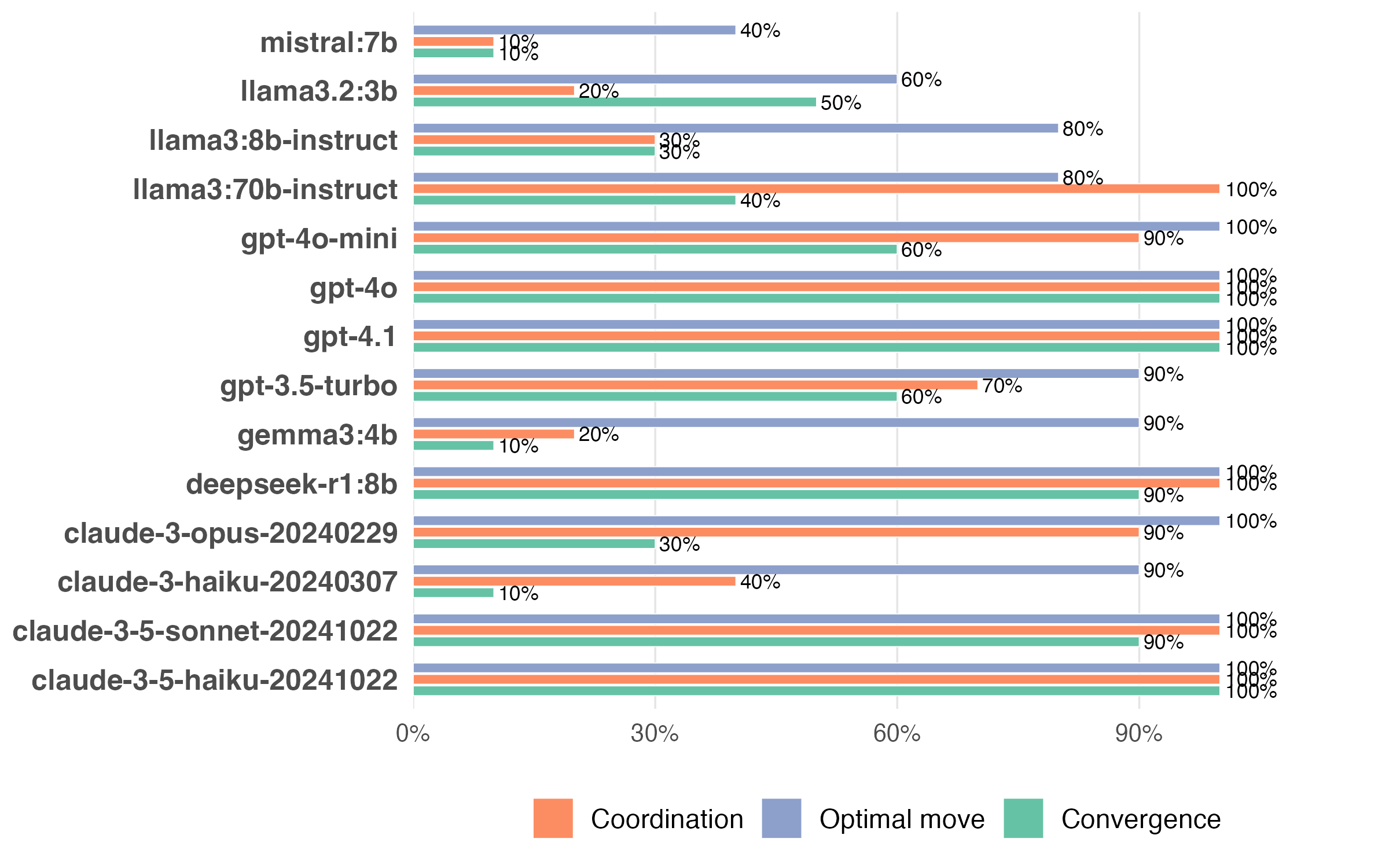}
  \caption{Percentage of runs per model where the model correctly identifies the dimension from the prompt.}
  \label{fig:annotations}
\end{figure}

We conducted this analysis across a range of different LLMs. We then also used the OpenAI model \texttt{gpt-4.1} to annotate three dimensions of the different LLM model outputs: whether it identified the setup as a coordination game; whether it correctly identified the optimal move; and whether it was able to correctly predict how the scenario would converge globally. We also asked the model to output the text snippet from the model output of the given LLM that the OpenAI model used to justify its decision.\footnote{We include the full prompt we used in the Appendix. The code itself we used can be found at the Github repo for the analyses we describe in this response at \url{https://github.com/cjbarrie/AI-norms-prompting/blob/main/03_annotate.py}} We then manually verify all of these annotations on the basis of the text snippet provided. We plot the results in Figure \ref{fig:annotations}. We see that for most models, including those used in the original paper, the models were able to correctly identify each of these dimensions. They are less able to predict the global convergence of the model---but many of the models manage this too. 

As a result, what we observe in the interaction of LLM agents may not be the ``spontaneous'' development of conventions as argued by the authors but the more prosaic mirroring of model training data. This is a problem that befalls not just the contribution by Ashery \textit{et al.} but arguably all simulations that rely on LLMs reproducing ``known'' emergent properties of complex human systems. Simply put, the emergent behaviors we believe we are seeing could just be the model reflecting its training data back at us.

So what would we need to see to believe that the LLMs were, in fact, self-organizing in ways similar to human populations? Modifying the model to hide the type of model through obfuscation -- say, for instance, by using unicode characters as tokens -- is unlikely to prevent the LLM from recognizing the general structure of the game with sufficient certainty. One solution would be to ``invent'' some completely novel game that the LLM has definitely not seen before. However, this requires us to come up with some novel model of human behavior that is of research interest, \textit{and} (assuming the model is correct) show that LLMs behave similarly. In generative ABMs, data leakage thus appears to be a fundamental and intractable issue without strict control over the training data. 

While there is an emerging literature on strategies for combating data leakage, this field is far from settled \citep{dong2024generalization}. Some promising avenues include interpretability probing with sparse autoencoders \citep{cunningham2023sparse} as well as the measurement of next-token perplexity based on a given prompt setup \citep{Li_Guerin_Lin_2024}. Whether or not the application of these techniques to generative ABM settings would help overcome some of the problems we describe above is an open question, however.

Overall, it remains a steep challenge to make claims of emergent behavior using existing off-the-shelf LLMs, in particular when using the models to reproduce or engage with existing research results. While conducting social simulations using LLMs remains a promising agenda, we believe the question of data contamination should be at the heart of future research in this area.

The question of data leakage in generative ABMs however also raises deeper philosophical questions. A long research tradition has shown that humans, too, are affected by research results. Most relevantly to the study at hand, research has found that economics students are more likely to reach equilibrium outcomes in coordination games \citep{carter1991economists}---a human form of ``data leakage''. Anthony Giddens \citep{giddens1993new} famously described this as a fundamental condition of the social sciences: unlike in the natural sciences, social science concepts can ``enter into'' and transform the very practices they aim to describe. Similarly, LLMs are shaped by the very act of scientific description, raising deep questions about their role in social scientific models. 

\bibliographystyle{plain}

\clearpage
\section{Appendix}

\begin{lstlisting}[frame=single,breaklines=true,basicstyle=\ttfamily\small]
System Prompt:

You are an expert annotation assistant. When given a model's response,
you must classify it along three dimensions and extract brief text
snippets to justify each classification. Respond ONLY with valid JSON
that matches the schema described in the user prompt.

User Prompt:

{response}

For each of the following, answer yes (1) or no (0), AND provide a short snippet
from the response that justifies your answer. Return JSON with these keys:
- coordination: did it identify this as a coordination game?
- coordination_justification: a brief excerpt supporting that answer
- optimal_move: did it state that the optimal move after a success is to keep answering the same way?
- optimal_move_justification: a brief excerpt supporting that answer
- convergence: did it predict that the game will converge to a unique global equilibrium?
- convergence_justification: a brief excerpt supporting that answer
\end{lstlisting}
\end{document}